\documentclass[runningheads]{llncs}
\usepackage{graphicx}
\usepackage{comment}
\usepackage{amsmath,amssymb} 
\usepackage{color}

\usepackage[width=122mm,left=12mm,paperwidth=146mm,height=193mm,top=12mm,paperheight=217mm]{geometry}

\usepackage{breqn}
\usepackage{float}
\usepackage{algorithm}
\usepackage{algorithmic}
\usepackage{capt-of}
\usepackage{pdfpages}
\usepackage{enumerate}
\usepackage{footnote}
\usepackage{multirow}
\usepackage{makecell}
\usepackage[utf8x]{inputenc}
\inputencoding{utf8}
\DeclareUnicodeCharacter{FF0C}{ }
\DeclareUnicodeCharacter{FF0D}{ }

\usepackage{booktabs}
\usepackage{multirow}
\usepackage{textcomp}
\usepackage[colorlinks,linkcolor=blue]{hyperref}

\begin{document}
\pagestyle{headings}
\mainmatter
\def\ECCVSubNumber{number 1399}  

\title{Bi-directional Cross-Modality Feature Propagation with Separation-and-Aggregation Gate for RGB-D Semantic Segmentation} 

\titlerunning{Bi-directional Cross-Modality Feature Propagation with SA-Gate} 
\authorrunning{Xiaokang Chen et al.} 
\author{	Xiaokang Chen \inst{1} \and
Kwan-Yee Lin \inst{2} \and
Jingbo Wang \inst{3} \and
Wayne Wu \inst{2}  \and \\
Chen Qian \inst{2}  \and
Hongsheng Li \inst{3} \and
Gang Zeng \inst{1} }
\institute{
$^1$Key Laboratory of Machine Perception (MOE), School of EECS, Peking University ~~\quad\\
\email{\{pkucxk,zeng\}@pku.edu.cn} \\
$^2$SenseTime Research \\
\email{\{linjunyi,wuwenyan,qianchen\}@sensetime.com} \\
$^3$The Chinese University of Hong Kong\\
\email{jbwang@ie.cuhk.edu.hk, hsli@ee.cuhk.edu.hk}
}
\maketitle

\begin{abstract}
Depth information has proven to be a useful cue in the semantic segmentation of RGB-D images for providing a geometric counterpart to the RGB representation. Most existing works simply assume that depth measurements are accurate and well-aligned with the RGB pixels and models the problem as a cross-modal feature fusion to obtain better feature representations to achieve more accurate segmentation. This, however, may not lead to satisfactory results as actual depth data are generally noisy, which might worsen the accuracy as the networks go deeper.

In this paper, we propose a unified and efficient Cross-modality Guided Encoder to not only effectively recalibrate RGB feature responses, but also to distill accurate depth information via multiple stages and aggregate the two recalibrated representations alternatively. The key of the proposed architecture is a novel Separation-and-Aggregation Gating operation that jointly filters and recalibrates both representations before cross-modality aggregation. Meanwhile, a Bi-direction Multi-step Propagation strategy is introduced, on the one hand, to help to propagate and fuse information between the two modalities, and on the other hand, to preserve their specificity along the long-term propagation process. Besides, our proposed encoder can be easily injected into the previous encoder-decoder structures to boost their performance on RGB-D semantic segmentation. Our model outperforms state-of-the-arts consistently on both in-door and out-door challenging datasets~{\footnote{Code of this work is available at {\color{blue} https://charlescxk.github.io/}}}.

\end{abstract}

\begin{figure*}[t!]
\centering
\includegraphics[width=1.0\textwidth]{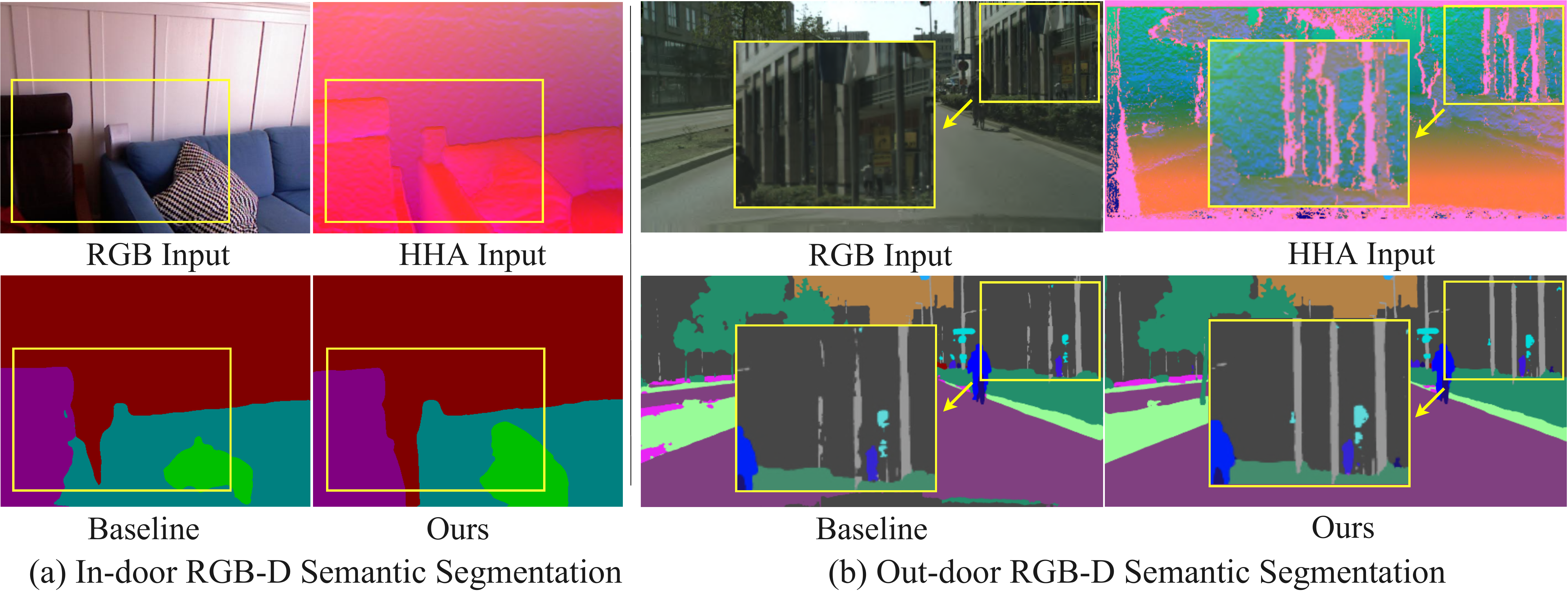}
\label{fig:intro}
\caption{\small{(a)RGB-D baseline, which is designed with a habitual cross-modality fusion schema, results in inaccurate classification on the area that exists substantial variations between RGB and Depth modalities. (b) The depth measurements in out-door environments are noisy. Without proposed modules, the results will degrade dramatically}}
\label{fig:intro-large}
\end{figure*}

\section{Introduction}

Semantic segmentation, which aims at assigning each pixel with different semantic labels, is a long-standing task. Besides exploiting various contextual information from the visual cues~\cite{fcn,danet,he2019dynamic,fu2019adaptive,spgnet,zhang2019acfnet}, depth data have recently been utilized as supplementary information to RGB data to achieve improved segmentation accuracy~\cite{park2017rdfnet,depthaware,pap,3dneighbourconv,he2017std2p,lstmcf,cheng2017locality,hung2019incorporating}.  Depth data naturally complements RGB signals by providing the 3D geometry to 2D visual information, which is robust to illumination changes and helps better distinguishing various objects.

Although significant advances have been achieved in RGB semantic segmentation, directly feeding the complementary depth data into existing RGB semantic segmentation frameworks~\cite{fcn} or simply ensemble results of two modalities~\cite{cheng2017locality} might lead to inferior performance.  The key challenges lie in two aspects. (1) \textit{The substantial variations between RGB and Depth modalities.} RGB and depth data show different characteristics. How to effectively identify their differences and unify the two types of information into an efficient representation for semantic segmentation is still an open problem. (2) \textit{The uncertainty of depth measurements.} Depth data provided with existing benchmarks are mainly captured by Time-of-Flight or structured light cameras, such as Kinect, AsusXtion and RealSense~\textit{etc.} The depth measurements are generally noisy due to different object materials and limited distance measurement range. The noise is more apparent for out-door scenes and results in undesirable segmentation, as shown in Fig~\ref{fig:intro-large}.

Most existing RGB-D based methods mainly focus on tackling the first challenge. Standard practice is to use the depth data~\footnote{Raw depth map or its encoded representation--HHA map, which includes horizontal disparity, height above ground and norm angle. For more detail about HHA, please refer to~\cite{hha}.} as another input and adopt Fully Convolutional Network (FCN)-like architectures with feature fusion schemas,~\textit{e.g.,} convolution and modality-based affinity~\textit{etc.,} to fuse the features of two modalities~\cite{park2017rdfnet,cheng2017locality,hu2019acnet,xing2019coupling}. The fused feature is then used to recalibrate the subsequent RGB feature responses or predicted results. Although these methods provide plausible solutions to unify the two types of information, the assumption of the input depth data being accurate and well-aligned with RGB signals might not be true, making these methods sensitive to in-the-wild samples. Moreover, how to ensure that the network fully utilizes information from both modalities remains an open problem. Recently, some works~\cite{pap,padnet} attempt to tackle the second challenge by diminishing the network's sensitivity to the quality of depth measurements. Instead of utilizing depth data as an extra input, they propose to distill the depth features via multi-task learning and regard depth data as extra supervision for training. Specifically,~\cite{padnet} introduces a two-stage framework, which first predicts several intermediate tasks including depth estimation and then uses the outputs of these intermediate tasks as the multi-modal input to final tasks.~\cite{pap} proposes a pattern-affinitive propagation with jointly predicting depth, surface normal and semantic segmentation to capture correlative information between modalities. We argue that there exists an inherent inefficacy in such design, \textit{i.e.} the interaction and correlation of RGB and depth information are only implicitly modeled. The complementarity of the two types of data for semantic segmentation was not well studied in this way.

Motivated by the above observations, we propose to tackle both two challenges in a simple yet effective framework by introducing a novel cross-modality guided encoder to FCN-like RGB-D semantic segmentation backbones. The key idea of the proposed framework is to leverage both channel-wise and spatial-wise correlation of the two modalities to firstly squeeze the exceptional feature responses of depth, which effectively suppresses feature responses from the low-quality depth measurements, and then use the suppressed depth representations to refine RGB features. In practice, we devise the steps bi-directionally due to the in-door RGB sources also contain noisy features. In contrast to depth data, the RGB noisy features are usually caused by similar appearance of different neighboring objects. We denote the above process as \textit{depth-feature recalibration} and \textit{RGB-feature recalibration}, respectively. We therefore introduce a new gate unit, namely the~\textit{Separation-and-Aggregation Gate (SA-Gate)}, to improve the quality of the multi-modality representation by encouraging the network to recalibrate and spotlight the modality-specific feature of each modality first, and then selectively aggregate the informative features from both modalities for the final segmentation.
To effectively take advantage of the differences of features between the two modalities, we further introduce the \textit{Bi-direction Multi-step Propagation (BMP)} that encourages the two streams to better preserve their specificity during the information interaction process in the encoder stage.

Our contributions can be summarized into three-fold:
\begin{itemize}

    \item We propose a novel bi-directional cross-modality guided encoder for RGB-D semantic segmentation. With the proposed \textit{SA-Gate} and \textit{BMP} modules, we could effectively diminish the influence of noisy depth measurements, and also allow incorporating sufficiently complementary information to form discriminative representations for segmentation.

    \item Comprehensive evaluation on the NYUD V2 dataset shows significant improvements by our approach when integrated into state-of-the-art RGB semantic segmentation networks, which demonstrate the generalization of our encoder as a plug-and-play module.
   
    \item The proposed method achieves state-of-the-art performances on both in-door and challenging out-door semantic segmentation datasets.
\end{itemize}

\section{Related Work}

\subsection{RGB-D Semantic Segmentation}

With the development of depth sensors, recently there is a surge of interest in leveraging
depth data as a geometry augmentation for RGB semantic segmentation task, dubbed as RGB-D semantic segmentation~\cite{park2017rdfnet,depthaware,kong2018recurrent,cfn,pap,chen20203d}. According to specific functionality of depth information suited in different architectures, current RGB-D based methods could be roughly divided into two categories.

Most of the works treat depth data as an additional input source to recalibrate the RGB feature responses either implicitly or explicitly. Long \textit{et al.}~\cite{fcn} shows simply averaging final score maps of RGB and D modalities helps enforce the inter-object discrimination in the in-door setting. Li \textit{et al.}~\cite{lstmcf} utilize the LSTM layers to selectively fuse the feature from the two modalities input. With a similar target, ~\cite{cheng2017locality} proposes locality-sensitive deconvolution networks along with a gated fusion module. Several recent works~\cite{DBLP:conf/eccv/WangWTSW16,deng2019rfbnet,hu2019acnet} extend the RGB feature recalibration process from the final outputs of a dual-path network to different stages of the backbone, encouraging better recalibration with multi-level cross-modality feature fusion. 
To guide the recalibration with explicit cross-modality interaction modeling, some works~\cite{kong2018recurrent,depthaware,3dgnn,xing20192} tailor general 2D operations to 2.5D behaviors with depth guidance. For example,~\cite{depthaware} proposes depth-aware convolution and pooling operations to help recalibrating RGB feature responses in depth-consistent regions. ~\cite{kong2018recurrent} proposes a depth-aware gate module that adaptively selects the pooling field size in a CNN according to object scale. 3DGNN~\cite{3dgnn} introduces a 3D graph neural network to model accurate context with geometry cues provided by depth.
Alternatively, some approaches regard the depth data as an extra supervised signal to recalibrate the RGB counterpart in a multi-task learning manner. For example, ~\cite{pap} proposes a pattern affinity propagation network to regularize and boost complementary tasks. ~\cite{padnet} introduces a multi-modal distillation model to pass the valid messages from depth to RGB features.

Different from previous works that hold the ideal assumption of depth source's quality and mainly focus on in-door setting, we try to extend the task to the in-the-wild environment, \textit{e.g.}, CityScapes dataset. The out-door setting is more challenging due to the inevitable noisy signals contained in the depth data. In this work, we try to recalibrate RGB feature responses from a filtered depth representation and vice versa, which effectively enhance the strength of representations for both modalities.

\subsection{Attention Mechanism}

Attention mechanisms have been widely utilized in kinds of computer vision tasks, serving as the tools to spotlight the most representative and informative regions of input signals~\cite{danet,woo2018cbam,wang2017residual,hu2018squeeze,sknet,non-local}.  For example, to improve the performance of the image/video classification task, SENet~\cite{hu2018squeeze} introduces a self－recalibrate gating mechanism by model importance among different channels of feature maps. Based on similar spirits, SKNet~\cite{sknet} designs a channel-wise attention module to select kernel sizes to adaptively adjust its receptive field size based on multiple scales of input information.~\cite{non-local} introduces a non-local operation which explores the similarity of each pair of points in space. For the segmentation task, a well-designed attention module could encourage the network to learn helpful context information effectively. For instance, DFN~\cite{dfn} introduces a channel attention block to select the more discriminative features from multi-level feature maps to get more accurate semantic information. DANet~\cite{danet} proposes two types of attention modules to model the semantic inter-dependencies in spatial and channel dimensions respectively. 

However, the main challenge of RGB-D semantic segmentation task is how to make full use of cross-modality data under the substantial variations and noisy signals between modalities. The proposed SA-Gate is the first to focus on the noisy features of cross-modalities by tailoring the attention mechanisms. The SA-Gate module is specialized for suppressing the exceptional noisy feature of depth data and recalibrate its counterpart RGB feature responses in a unified manner at first, and then fuses the cross-modality information with a softmax gating that is guided by the recalibrated features, achieving effective and efficient cross-modality feature aggregation.

\begin{figure*}[t!]
\centering
\includegraphics[width=0.8\textwidth]{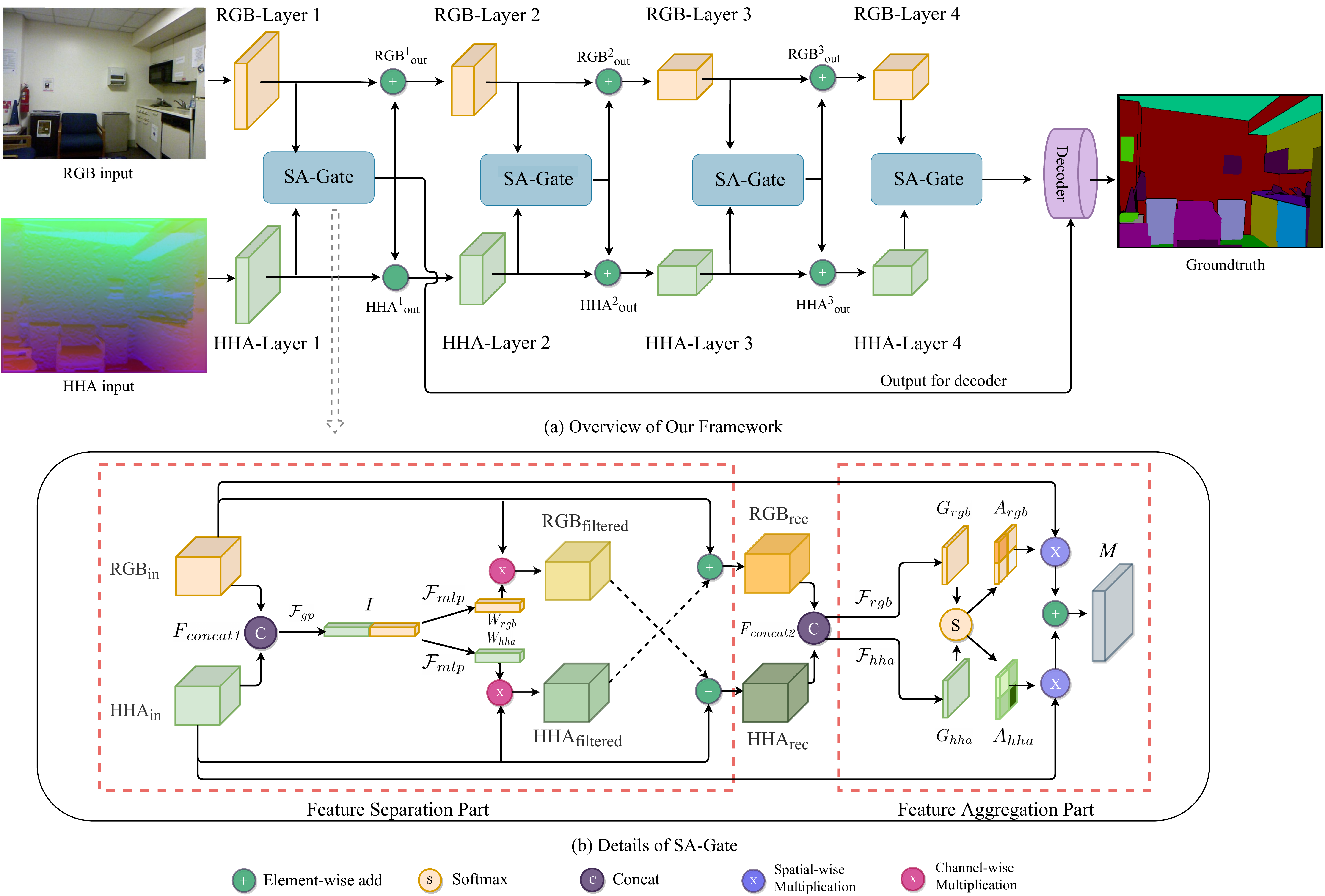}
\caption{\small{(a)The overview of our network. We employ an encoder-decoder architecture. The input of the network is a pair of RGB-HHA images. During training, each pair of feature maps (\textit{e.g.}, outputs of RGB-Layer1 and HHA-Layer1) are fused by a SA-Gate and propagated to the next stage of the encoder for further feature transformation. Fusion results of the first and the last SA-Gates would be propagated to the segmentation decoder (DeepLab V3+). (b) The architecture of the SA-Gate, which contains two parts, Feature Separation (FS) and Feature Aggregation (FA)}}

\label{fig:whole-arch}
\end{figure*}

\section{Method}

RGB-D semantic segmentation needs to aggregate features from both RGB and depth modalities. However, both modalities have inevitably noisy information.
Specifically, depth measurements are inaccurate due to the characteristics of depth sensors and RGB features might generate confusing results due to the high appearance similarity between the objects. An effective cross-modality aggregation scheme should be able to identify their strengths from each feature as well as unify the most informative cross-modality features into an efficient representation.  To this end, we put forward a novel cross-modality guided encoder. The overall framework of the proposed approach is depicted in Fig. \ref{fig:whole-arch} (a), which consists of a cross-modality guided encoder and a segmentation decoder. Given RGB-D data as inputs~\footnote{Note that we use HHA map to encode the depth measurements.}, our encoder recalibrates and fuses the complementary information from the two modalities via the SA-Gate unit, and then propagates the fused multi-modal features along with modality-specific features via the Bi-direction Multi-step Propagation (BMP) module. The information is then decoded by a segmentation decoder network to generate the segmentation map. We will detail each component in the remaining parts of this section.

\subsection{Bi-direction Guided Encoder}

\noindent \textbf{Separation-and-Aggregation (SA) Gate.} To ensure informative feature propagation between modalities, the SA-Gate is designed with two operations. One is feature recalibration on each single modality, and the other is cross-modality feature aggregation. The operations are in terms of Feature Separation (FS) and Feature Aggregation (FA) parts, as illustrated in Fig~\ref{fig:whole-arch} (b).

\begin{figure}[h]
\centering
\includegraphics[width=0.9\textwidth]{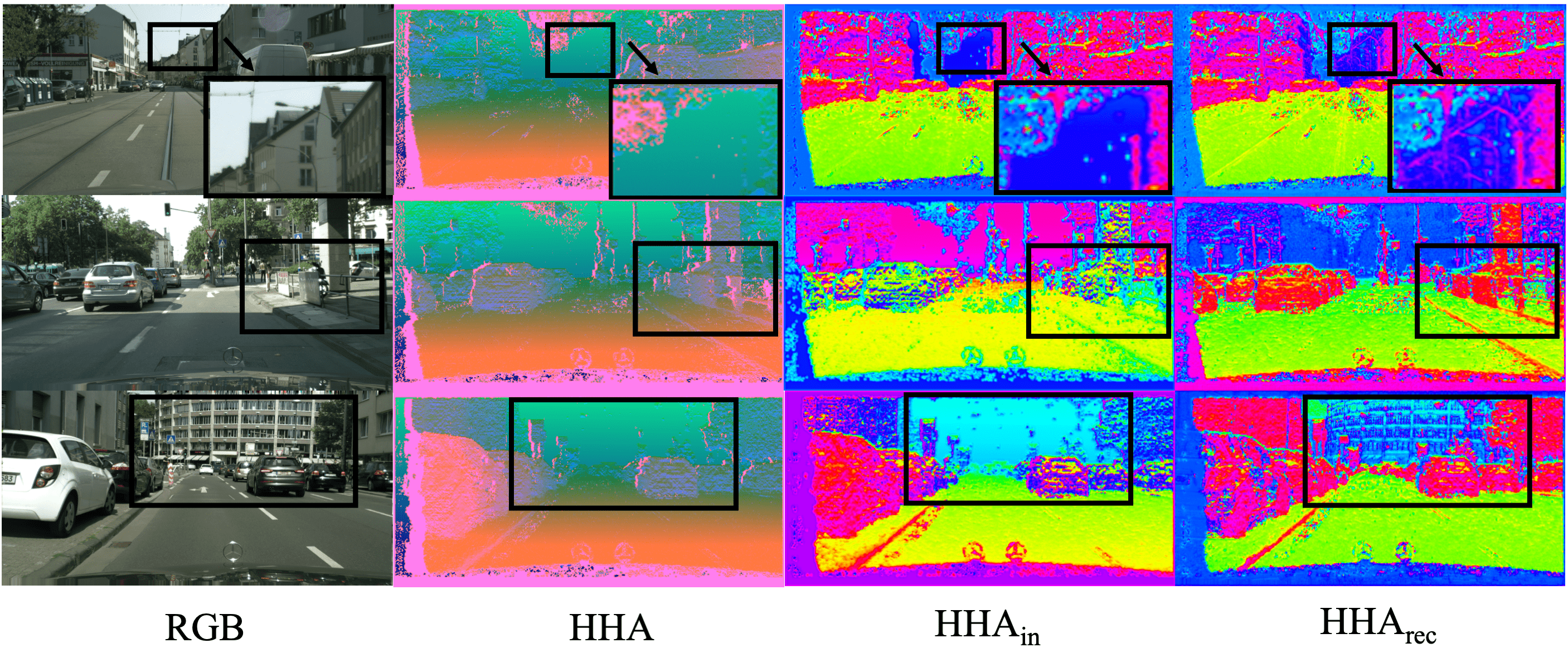} 
\caption{Visualization of depth features before and after FSP on CityScapes validation set. We can observe that objects have more precise shapes after FSP and invalid partial surfaces are completed. More explanation is illustrated in the supplemental material}
\label{denoise-city-hha}
\end{figure}

\textit{Feature Separation (FS)}. We take depth stream for example. Due to physical characteristics of depth sensors, noisy signals in depth modality frequently show up in regions close to object's boundaries or partial surfaces outside the scope of depth sensors, as shown in the second column of Fig.~\ref{denoise-city-hha}. Hence, the network is expected to first filter noisy signals surrounding these local regions to avoid misleading information propagation on the process of recalibrating complementary RGB modality and aggregating cross-modality features. In practice, we exploit high confident activations in RGB stream to filter out exceptional depth activations at the same level. To do so, global spatial information of both modalities should be embedded and squeezed to obtain a cross-modality attention vector first. We achieve this by a global average pooling along the channel-wise dimensions of two modalities, which is followed by concatenation and a MLP operation to obtain attention vector. Suppose we have two input feature maps denoted as  $\rm{RGB}_{in} \in \mathbb{R}^{C \times H \times W}$ and $\rm{HHA}_{in} \in \mathbb{R}^{C \times H \times W}$, above operations could be formulated as
\begin{equation}
   I = \mathcal{F}_{gp}(\rm{RGB}_{in}\parallel\rm{HHA}_{in}),
\end{equation}
where $\parallel$ denotes the concatenation of feature maps from two modalities, $\mathcal{F}_{gp}$ refers to global average pooling, $I=(I_1,\dots,I_k,\dots, I_{2C})$ is the cross-modality global descriptor for collecting expressive statistics for the whole inputs. Then, the cross-modality attention vector for the depth input is learned by
\begin{equation}
W_{hha} = \sigma(\mathcal{F}_{mlp}(I)), \quad W_{hha} \in \mathbb{R}^{C},
\end{equation}
where $\mathcal{F}_{mlp}$ denotes MLP network, $\sigma$ denotes sigmoid function scaling the weight value into $(0, 1)$. By doing so, the network can take advantage of the most informative visual appearance and geometry features, and thus tends to effectively suppress the importance of noisy features in depth stream. Then, we could obtain a less noisy depth representation, namely Filtered HHA, through a channel-wise multiplication $\circledast$ between input depth feature maps and the cross-modality gate:
\begin{equation}
\rm{HHA}_{filtered} = \rm{HHA}_{in}\circledast W_{hha}.
\end{equation}

With a filtered depth representation counterpart, the RGB feature responses could be recalibrated with more accurate depth information. We devise the recalibration operation as the summation of the two modalities:
\begin{equation}
    \rm{RGB_{rec}} =\rm{HHA}_{filtered}+ \rm{RGB_{in}},
    \label{sum_cross}
\end{equation}
where 
$\rm{RGB_{rec}}$ denotes recalibrated RGB feature maps. The general idea behind the formula is that, instead of directly using element-wise product to reweight RGB feature with regarding depth features as recalibrate coefficients, the proposed operation using summation could be viewed as some kind of offset to refine RGB feature responses at corresponding positions, as demonstrated in Table \ref{FS-ab}.

In practice, we implement \textit{recalibration step} in a symmetric and bi-directional manner, such that low confident activations in RGB stream could also be suppressed in the same manner and filtered RGB information $\rm{RGB_{filtered}}$ could inversely recalibrate the depth feature responses to form a more robust depth representation $\rm{HHA_{rec}}$. We visualize feature maps of HHA before and after Feature Separation Part in Fig.~\ref{denoise-city-hha}. The RGB counterpart is shown in the supplementation.

\textit{Feature Aggregation (FA)}. RGB and D features are strongly complementary to each other. To make full use of their complementarity, we need to complementarily aggregate the cross-modality features at a certain position in space according to their characterization capabilities. To achieve this, we consider both characteristics of these two modalities and generate spatial-wise gates for both $\rm{RGB}_{in}$ and $\rm{HHA}_{in}$ to control information flow of each modality feature map with soft attention mechanism, which is visualized in Figure \ref{fig:whole-arch} (b) and marked by the second red frame.  To make the gate more precise, we use recalibrated RGB and HHA feature maps from \textit{FS} part, \textit{i.e.}, $\rm{RGB}_{rec} \in \mathbb{R}^{C \times H \times W}$ and $\rm{HHA}_{rec} \in \mathbb{R}^{C \times H \times W}$, to generate the gate. We first concatenate these two feature maps to combine their features at a certain position in space. Then we define two mapping functions to map high-dimensional feature to two different spatial-wise gates:
\begin{align}
\mathcal{F}_{rgb}: &\mathit{F_{concat2}} \rightarrow G_{rgb} \in \mathbb{R}^{1 \times H \times W} \label{mapping1},\\
\mathcal{F}_{hha}: &\mathit{F_{concat2}} \rightarrow G_{hha} \in \mathbb{R}^{1 \times H \times W} \label{mapping2},
\end{align}
where $\mathit{F_{concat2}} \in \mathbb{R}^{2C \times H \times W}$ is the concatenated feature, $G_{rgb}$ is the spatial-wise gate for RGB feature map, and $G_{hha}$ is the spatial-wise gate for HHA feature map. In practice, we use a $1\times 1$ convolution to implement this mapping function. A softmax function is applied on these two gates:
\begin{equation}
A_{rgb}^{(i,j)} = \frac{e^{G_{rgb}^{(i,j)}}}{e^{G_{rgb}^{(i,j)}} + e^{G_{hha}^{(i,j)}}}, \,\, A_{hha}^{(i,j)} = \frac{e^{G_{hha}^{(i,j)}}}{e^{{G_{rgb}^{(i,j)}}} + e^{G_{hha}^{(i,j)}}}，
\end{equation}
where $A_{rgb}, A_{hha} \in  R^{1 \times H \times W}$ and $A_{rgb}^{(i,j)} + A_{hha}^{(i,j)} = 1$. $G_{rgb}^{(i,j)}$ is the weight assigned to each position in the RGB feature map and $G_{hha}^{(i,j)}$ is the weight assigned to each position in the HHA feature map. The final merged feature $M$ can be obtained by weighting the RGB and HHA maps:
\begin{equation}
M_{i,j} = {\rm RGB}_{in}^{(i,j)} \cdot A_{rgb}^{(i,j)} + {\rm HHA}_{in}^{(i,j)} \cdot A_{hha}^{(i,j)}.
\end{equation}

So far, we have added gated RGB and HHA feature maps to obtain the fused feature maps $M$. Since SA-Gate is injected into the encoder stage, we then average the fused features and the original input to obtain $\rm{RGB}_{out}$ and $\rm{HHA}_{out}$ respectively, which share similar spirits with residual learning.

\noindent \textbf{Bi-directional Multi-step Propagation (BMP).}
By normalizing the sum of two weights at each position to $1$, the numerical scale of the weighted feature will not significantly differ from the input RGB or HHA. Therefore, it has no negative influence on the learning of the encoder or the loading of the pre-trained parameters. For each layer $l$, we use the output $ M^l$ generated by the $l$-th SA-Gate to refine the raw output of the $l$-th layer in the encoder: ${\rm RGB}_{out}^l  = ({\rm RGB}_{in}^l + M^l) / 2$, ${\rm HHA}_{out}^l = ({\rm HHA}_{in}^l + M^l) / 2$. This is a bi-directional propagation process and the refined results will be propagated to the next layer in the encoder for more accurate and efficient encoding of the two modalities.

\subsection{Segmentation Decoder}
The decoder can adopt almost any design of decoder from SOTA RGB-based segmentation networks, since SA-Gate is a plug-and-play module and can make good use of complementary information of cross-modality on encoder stage. We show results of combining our encoder with different decoders in Table \ref{different-decoders}. We choose DeepLabV3+~\cite{v3+} as our decoder for it achieves the best performance.

\section{Experiments}
We conduct comprehensive experiments on in-door NYU Depth V2 and out-door CityScapes datasets in terms of two metrics: mean Intersection-over-Union ($mIoU$) and pixel accuracy (pixel acc.). We also evaluate our model on SUN-RGBD dataset (Please refer to the supplemental material for more details).

\subsection{Datasets}
\noindent \textbf{NYU Depth V2}~\cite{nyudv2} contains $1449$ RGB-D images with 40-class labels, in which $795$ images are used for training and the rest 654 images are for testing. 

\noindent \textbf{CityScapes}~\cite{cordts2016cityscapes} contains images from $27$ cities. There are $2975$ images for training, $500$ for validation and $1525$ for testing. Each image has a resolution of $2048 \times 1024$ and is fine-annotated with pixel-level labels of $19$ semantic classes. \textbf{We do not use additional coarse annotations in our experiments}.

\subsection{Implementation Details}
We use PyTorch framework. For data augmentation, we use random horizontal flipping and scaling with scales $[0.5$,$1.75]$. When comparing with SOTA methods, we adopt flipping and multi-scale inference strategies as a test-time augmentation to boost the performance. More details are shown in the supplemental material.

\begin{table}
\begin{center}
\caption{Comparison of efficiency on NYUDV2 test set. We use ResNet-50 as backbone and DeepLab V3+\cite{v3+} as decoder. FLOPS are estimated for input of $3 \times 480 \times 480$}
\label{efficiency-ab}
\resizebox{0.66\columnwidth}{!}{
\setlength{\tabcolsep}{2mm}{
\begin{tabular}{cccc}
\toprule
\textbf{Methods} & \textbf{Params/M}  & \textbf{FLOPs/G} & \textbf{mIoU(\%)} \\
\hline
RGB-D baseline & 78.2 & 269.6 & 46.7 \\
Ours & \textbf{63.4} & \textbf{204.9} & \textbf{50.4} \\
\bottomrule
\end{tabular}}
}
\end{center}
\end{table}

\subsection{Efficiency Analysis}
To verify whether the proposed cross-modality feature propagation helps and is efficient, we compare the final model with the RGB-D baseline. We average predictions of two parallel DeepLab V3+ as RGB-D baseline. As shown in Table \ref{efficiency-ab}, the proposed method achieves better performance with significantly less memory requirement and computational cost when compared with baseline. The results indicate that aimlessly adding parameters to a multi-modality network will not bring extra representational power to better recognize objects. In contrast, a well-design cross-modality mechanism, like proposed cross-modality feature propagation, helps to learn more powerful representations to improve performance more efficiently.

\begin{table}
\begin{center}
\caption{Ablation study on \textit{feature separation (FS)} part on NYU Depth V2 test set. No decoder is used here}
\label{FS-ab}
\resizebox{0.8\columnwidth}{!}{
\setlength{\tabcolsep}{2mm}{
\begin{tabular}{ccccccc}
\toprule
\textbf{Backbone} & \textbf{Concat}  & \textbf{Self-global} & \textbf{Cross-global} & \textbf{Product} & \textbf{Proposed} & \textbf{mIoU(\%)} \\
\hline
Res50 & \checkmark & & & & & 47.8\\
Res50 &  &  \checkmark & & & & 47.5\\
Res50 &  &  & \checkmark & & & 47.8\\
Res50 &  &  &  & \checkmark & & 47.5\\
Res50 & & & & & \checkmark & \textbf{48.6}\\
\bottomrule
\end{tabular}}
}
\end{center}
\end{table}


\subsection{Ablation Study}
\label{sec-abstudy}
We perform ablation studies on our design choices under same hyperparameters.

\noindent \textbf{Feature Separation.} We employ the FS operation before the feature aggregation in SA-Gate, to filter out noisy features for bi-directional recalibration step. To verify effectiveness of this operation, we ablate each design of FS in Table \ref{FS-ab}. Note that we ablate four different architectures and replace all FS parts in the network for comparison. `{Concat}' represents we concatenate $\rm{RGB}_{in}$ and $\rm{HHA}_{in}$ feature maps and directly pass them to feature aggregation part. `{Self-global}' represents we filter single modality features with its own global information. `{Cross-global}' represents the filtered RGB is added to input RGB and vice versa. The filtering guidance comes from cross-modality global information.
`{Product}' means we multiply $\rm{RGB}_{in}$ by $\rm{HHA}_{filtered}$ and vice versa. We see that from column $2$ to $4$, not using cross-modality information to filter noisy feature or refine features without explicit cross-modality recalibration lead to about $1\%$ drop. On the other hand, the last two columns indicate the cross-modality guidance (E.q~\ref{sum_cross}) is more appropriate and effective than cross-modality re-weighting when doing cross-modality recalibration. Overall, these results show that proposed FS operator effectively filters incorrect messages and recalibrates feature responses, achieving the best performance among all compared designs.


\begin{table}[t]
\begin{center}
\caption{Ablation study on \textit{feature aggregation (FA)} part on NYU Depth V2 test set. No decoder is used here}
\label{FA-ab}
\resizebox{0.6\columnwidth}{!}{
\setlength{\tabcolsep}{2mm}{
\begin{tabular}{ccccc}
\toprule
\textbf{Backbone} & \textbf{Addition}  & \textbf{Conv} & \textbf{Proposed} & \textbf{mIoU(\%)} \\
\hline
Res50 & \checkmark & & &  47.8 \\
Res50 &  &  \checkmark & &  48.0 \\
Res50 & & & \checkmark & \textbf{48.6}\\
\bottomrule
\end{tabular}}
}
\end{center}
\end{table}

\noindent \textbf{Feature Aggregation.} We employ the SA-Gating mechanism to adaptively select the feature from the cross-modal data, according to their different characteristics at each spatial location. This gate can effectively control information flow of multimodal data. To evaluate the validity of the design, we perform ablation study on feature aggregation, as shown in Table \ref{FA-ab}. The experiment setting is kept the same as above.  `{Addition}' represents directly adding the recalibrated RGB and HHA feature maps. `{Conv}' represents conducting convolution on the concatenated feature map. `{Proposed}' represents the FA operator. We see that FA operator leads to the best result, since it considers the spatial-wise relationship between two modalities and can better explore the complementary information.

\begin{table}[t]
\begin{center}
\caption{Ablation study on encoder design on NYU Depth V2 test set. '*' means we average two outputs of RGB and HHA to get final output. No decoder is used here}
\label{stage-ab}
\resizebox{0.66\columnwidth}{!}{
\setlength{\tabcolsep}{2mm}{
\begin{tabular}{cccccc}
\toprule
\textbf{Backbone} & \textbf{Block1} & \textbf{Block2}  & \textbf{Block3} &  \textbf{Block4} & \textbf{mIoU(\%)} \\
\hline
Res50$^*$ & & & & & 45.9 \\
Res50$^*$ & \checkmark & & & & 47.8 \\
Res50$^*$ & & \checkmark & & & 47.5 \\
Res50$^*$ & & & \checkmark & & 46.8 \\
Res50$^*$ & & & & \checkmark & 44.3 \\
Res50$^*$ & \checkmark & \checkmark & & & 47.9 \\
Res50$^*$ & \checkmark & \checkmark & \checkmark & & 48.3 \\
Res50$^*$ & \checkmark & \checkmark & \checkmark & \checkmark & 48.0\\
\noindent Res50$^{\,\,\,}$ & \checkmark & \checkmark & \checkmark & \checkmark & \textbf{48.6}\\
\bottomrule
\end{tabular}}
}
\end{center}
\end{table}

\begin{table}[t]
\begin{center}
\caption{Ablation study for BMP and SA-Gate. No decoder is used here}
\label{prop-ab}
\resizebox{0.5\columnwidth}{!}{
\setlength{\tabcolsep}{3mm}{
\begin{tabular}{lc}
\toprule
\textbf{Method} & \textbf{mIoU(\%)} \\
\hline
Res50 (Average of Dual Path) & 45.9 \\
Res50 + SA-Gate & 47.4 $(1.5 \%\uparrow)$ \\
Res50 + BMP & 47.8 $(1.9\% \uparrow)$\\
Res50 + BMP + SA-Gate & \textbf{48.6} $(2.7\% \uparrow)$ \\
\bottomrule
\end{tabular}}
}
\end{center}
\end{table}

\begin{table}[t]
\begin{center}
\caption{\small{The plug-and-play property evaluation of the proposed model on NYU Depth V2 test set. \textbf{Method} indicates different decoders, SA-Gate indicates the proposed fusion module. \textbf{RGB}: RGB image as inputs; \textbf{RGB-D}: the simple method which only average final score maps of RGB path and HHA path. Note that we reproduce these methods using official open-source code and all experiments use the same setting as our method}}
\label{different-decoders}
\resizebox{0.9\columnwidth}{!}{
\setlength{\tabcolsep}{2mm}{
\begin{tabular}{lccc}
\toprule
\textbf{Method}  & \textbf{RGB(\%\emph{mIoU})} & \textbf{RGB-D(\%\emph{mIoU})} & \textbf{RGB-D w SA-Gate(\%\emph{mIoU})} \\
\hline
DeepLab V3~\cite{v3} & 44.7 & 46.5 & \textbf{49.1 $(2.6 \uparrow)$}   \\
PSPNet~\cite{pspnet} & 43.1 & 46.2 & \textbf{48.2 $(2.0 \uparrow)$}\\
DenseASPP~\cite{yang2018denseaspp} & 42.3 & 45.7 & \textbf{47.8 $(2.1 \uparrow)$} \\
OCNet~\cite{yuan2018ocnet} & 44.5 & 47.6 & \textbf{49.1 $(1.5 \uparrow)$} \\
DeepLab V3+~\cite{v3+} & 44.3 & 46.7 &\textbf{50.4 $(3.7 \uparrow)$} \\
DANet~\cite{danet} & 43.0 & 45.5 & \textbf{48.6 $(3.1 \uparrow)$} \\
FastFCN~\cite{wu2019fastfcn} & 45.4 & 47.6 & \textbf{50.1 $(2.5 \uparrow)$}\\
\bottomrule
\end{tabular}}
}
\end{center}
\end{table}
\begin{figure*}[t]
\centering
\includegraphics[width=1\textwidth]{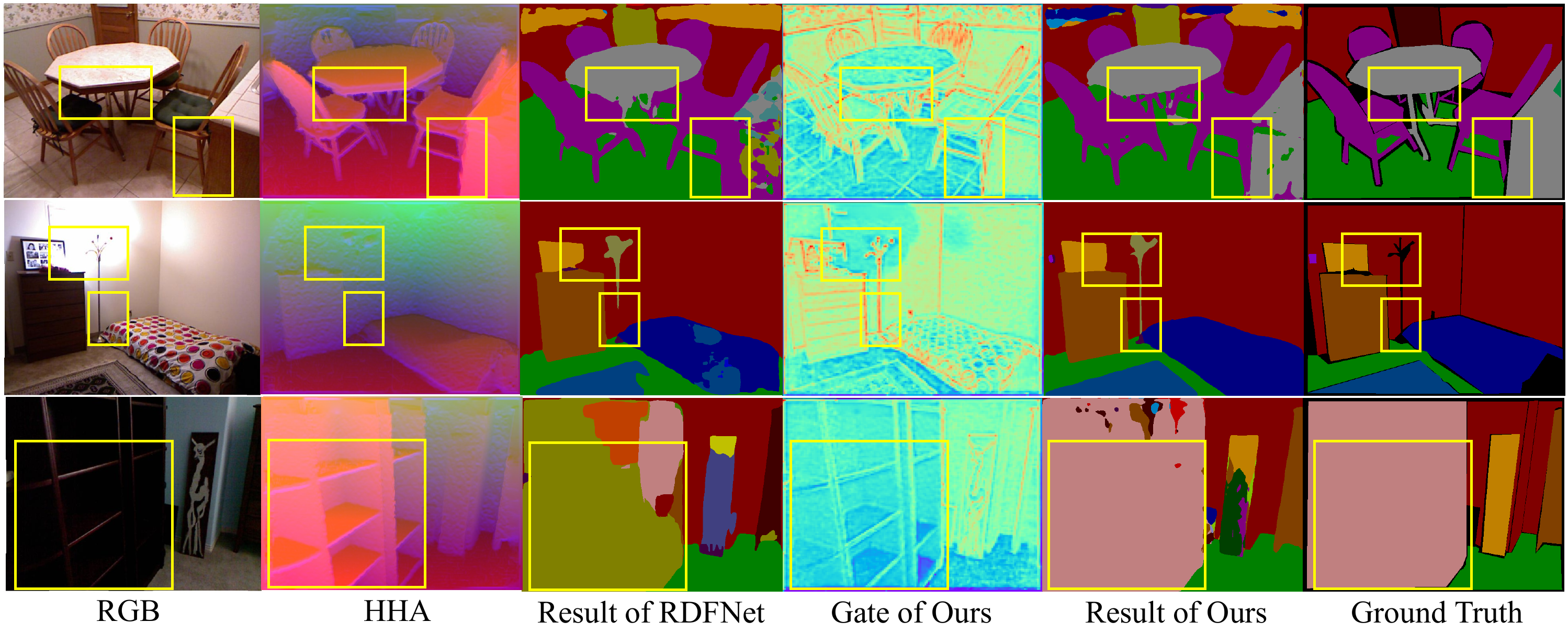}
\caption{\small{Visualization of feature selection through SA-Gate on NYUD V2 test set. For each row, we show (1) RGB, (2) HHA, (3) results of RDFNet-101, (4) visualization of SA-Gate, (5) results of ours, (6) GT. Red represents a higher weight assigned to RGB and blue represents a higher weight assigned to HHA. Best viewed in color}}
\label{fig:different-fuse}
\end{figure*}

\begin{table}[t]
\begin{center}
\caption{\small{State-of-the-art comparison experiments on NYU Depth V2 test set}}
\label{sota-nyu}
\resizebox{0.5\columnwidth}{!}{
\setlength{\tabcolsep}{2mm}{
\begin{tabular}{lcc}
\toprule
\textbf{Method} & \textbf{mIoU(\%) } & \textbf{Pixel Acc.(\%)} \\
\hline
3DGNN~\cite{3dgnn} & 43.1 & - \\
Kong $et\ al.$~\cite{kong2018recurrent} & 44.5 & 72.1 \\
LS-DeconvNet~\cite{cheng2017locality} & 45.9 & 71.9\\
CFN~\cite{cfn} & 47.7 & - \\
ACNet~\cite{hu2019acnet} & 48.3 & - \\
RDF-101~\cite{park2017rdfnet} & 49.1 & 75.6\\
PADNet~\cite{padnet} & 50.2 & 75.2 \\
PAP~\cite{pap} & 50.4 & 76.2\\
\hline
Ours & \textbf{52.4} & \textbf{77.9} \\
\bottomrule
\end{tabular}}
}
\end{center}
\end{table}
\setlength{\tabcolsep}{1.4pt}

\begin{table}[t]
\begin{center}
\caption{\small{Cityscapes test set accuracies. `*' means RGB-D based methods}}
\label{city-test}
\resizebox{\columnwidth}{!}{
\begin{tabular}{l|lllllllllllllllllll|c}
\toprule
Method &  \rotatebox{90}{roa.} &  \rotatebox{90}{sid.} &  \rotatebox{90}{bui.} &  \rotatebox{90}{wal.} &  \rotatebox{90}{fen.} &  \rotatebox{90}{pol.} &  \rotatebox{90}{lig.} &  \rotatebox{90}{sig.} &  \rotatebox{90}{veg.} &  \rotatebox{90}{ter.} &  \rotatebox{90}{sky} &  \rotatebox{90}{per.} &  \rotatebox{90}{rid.} &  \rotatebox{90}{car} &  \rotatebox{90}{tru.} &  \rotatebox{90}{bus} &  \rotatebox{90}{tra.} &  \rotatebox{90}{mot.} &  \rotatebox{90}{bic.} & mIoU \\

\hline
 DUC~\cite{duc} & 98.6 & 86.1 & 93.5 & 56.1 & 63.3 & 69.7 & 77.3 & 81.3 & 93.9 & 72.9 & 95.7 & 87.3 & 72.9 & 96.2 & 76.8 & 89.4 & 86.5 & 72.2 & 78.2 & 77.6\\
 DenseASPP~\cite{yang2018denseaspp} & 98.7 & 87.1 & 93.4 & 60.7 & 62.7 & 65.6 & 74.6 & 78.5 & 93.6 & 72.5 & 95.4 & 86.2 & 71.9 & 96.0 & 78.0 & 90.3 & 80.7 & 69.7 & 76.8 & 80.6\\
CCNet~\cite{huang2018ccnet} & - & - & - & - & - & - & - & - & - & - & - & - & - & - & - & - & - & - & - & 81.4\\
BFP~\cite{bfp} & 98.7 & 87.0 & 93.5 & 59.8 & 63.4 & 68.9 & 76.8 & 80.9 & 93.7 & 72.8 & 95.5 & 87.0 & 72.1 & 96.0 & 77.6 & 89.0 & 86.9 & 69.2 & 77.6 & 81.4\\
DANet~\cite{danet} & 98.6 & 86.1 & 93.5 & 56.1 & 63.3 & 69.7 & 77.3 & 81.3 & 93.9 & 72.9 & 95.7 & 87.3 & 72.9 & 96.2 & 76.8 & 89.4 & 86.5 & 72.2 & 78.2 & 81.5\\
 GALD~\cite{li2019global} & 98.7 & 87.2 & 93.8 & 59.3 & 61.9 & 71.4 & 79.2 & 82.0 & 93.9 & 72.8 & 95.6 & 88.4 & 74.8 & 96.3 & 74.1 & 90.6 & 81.1 & 73.4 & 79.8 & 81.8 \\
ACFNet~\cite{zhang2019acfnet} & 98.7 & 87.1 & 93.9 & 60.2 & 63.9 & 71.1 & 78.6 & 81.5 & 94.0 & 72.9 & 95.9 & 88.1 & 74.1 & 96.5 & 76.6 & 89.3 & 81.5 & 72.1 & 79.2 & 81.8 \\
\hline
LDFNet$^*$~\cite{hung2019incorporating} & - & - & - & - & - & - & - & - & - & - & - & - & - & - & - & - & - & - & - & 71.3\\
Shu Kong $et\ al.$$^*$~\cite{kong2018recurrent} & - & - & - & - & - & - & - & - & - & - & - & - & - & - & - & - & - & - & - & 78.2\\
PADNet $^*$~\cite{padnet} & - & - & - & - & - & - & - & - & - & - & - & - & - & - & - & - & - & - & - & 80.3\\
Choi $et\ al.$$^*$~\cite{choi2020cars} & 98.8 & 88.0 & 93.9 & 60.5 & 63.3 & 71.3 & 78.1 & 81.3 & 94.0 & 72.9 & 96.1 & 87.9 & 74.5 & 96.5 & 77.0 & 88.0 & 85.9 & 72.7 & 79.0 & 82.1\\

\hline
RGB baseline (Deeplab V3+~\cite{v3+}) & 98.7 & 87.1 & 93.9 & 61.0 & 63.8 & 71.5 & 78.6 & 82.6 & 93.9 & 72.6 & 95.9 & 88.3 & 74.8 & 96.5 & 68.9 & 86.1 & 86.4 & 73.6 & 79.1 & 81.8\\
RGB-D baseline$^*$ & 98.7 & 86.7 & 93.7 & 57.8 & 61.8 & 70.0 & 77.3 & 81.8 & 93.9 & 72.2 & 95.9 & 87.9 & 74.1 & 96.3 & 70.7 & 87.9 & 80.3 & 72.2 & 78.6 & 80.9\\
Ours$^*$ & 98.7 &	87.3 &	93.9 &	63.8 &	62.7 &	70.8 &	77.9 &	82.2 &	93.9 &	72.8 &	95.9 &	88.2 &	75.2 &	96.5 &	80.4 &	91.6 &	89.0 &	73.2 &	78.9  & \textbf{82.8}\\
\bottomrule
\end{tabular}
}
\end{center}
\end{table}

\noindent \textbf{Design of Encoder.}
We verify and analyze the effectiveness of proposed BMP to our encoder, and how it functions with the SA-Gate. Toward this end, we conduct two ablation studies as shown in Table~\ref{stage-ab} $\&$~\ref{prop-ab}. We use ResNet-50 as our backbone here and directly upsampling the final score map by a factor of 16, without using a segmentation decoder. The first row 
in Table~\ref{stage-ab} $\&$~\ref{prop-ab} is the baseline that averages score maps generated by two ResNet-50 (RGB \& D). 

For the first ablation, we gradually embed SA-Gate unit behind different layers of ResNet50. Note that we generate score maps for both two sides and average them as final segmentation result. This setting is different from those above, because last block of ResNet may not be equipped with a SA-Gate in this part, \textit{i.e.}, no fused feature is generated from last block. From Table \ref{stage-ab}, we observe that if SA-Gate is embedded into a higher stage, it will lead to relatively worse performance.  Besides, when stacking SA-Gate stage by stage, the additional gain continuously reduces. These two phenomena show that features of different modalities are more different in lower stage and an early fusing will achieve better performance. Table~\ref{prop-ab} shows results of second experiment. We observe that both SA-Gate and BMP can boost performance. Meanwhile, they complement each other and performs better in the presence of the other component. Moreover, when associating Table~\ref{prop-ab} $\&$~\ref{FS-ab}, we see that SA-Gate helps BMP better propagate valid information than other gate mechanisms. It demonstrates effectiveness and importance of a more accurate representation to the feature propagation.

\noindent \textbf{The Plug-and-Play Property of Proposed Encoder.} We conduct ablation study to validate the flexibility and effectiveness of our method for different types of decoders. Following recent RGB-based semantic segmentation algorithms, we splice their decoders with our model to form modified RGB-D versions (\textit{i.e.,} RGB-D w SA-Gate), as shown in Table~\ref{different-decoders}. We see that in the column $2$ and $4$, our method consistently helps
achieving significant improvements against original RGB versions.
Besides, comparing with naive RGB-D modifications, our method also boosts the performance at least 1.5\% \emph{mIoU}. Especially, with the decoders in Deeplab V3+~\cite{v3+}, our method achieves 3.7\% \emph{mIoU} improvements. The results verify both the flexibility and effectiveness of our method for various decoders.

\subsection{Visualization of SA-Gate}
We visualize first SA-Gate in our model to see what it has learned, as shown in Fig \ref{fig:different-fuse}.
~\textit{ Note that the black region in GT represents ignored pixels when calculating IoU. We reproduce RDFNet-101~\cite{park2017rdfnet} in PyTorch with 48.7\% mIoU on NYU Depth V2, which is close to the result in the original paper (49.1\%)}. Red represents a higher weight assigned to RGB and blue represents a higher weight assigned to HHA. From column 4, we can see that RGB has a stronger response at boundary and HHA responds well in glare and dark areas. The phenomenon is reasonable since RGB feature has more details in high contrast areas and HHA feature is not affected by lighting conditions. From row 1, details inside yellow boxes are lost in HHA while obvious in RGB. Our method successfully identifies chair legs and distinguishes table that looks similar to chair. In row 2, glare blurs the border of the photo frame. 
Since our model focuses more on HHA in this area, it predicts the photo frame more completely than RDFNet. Besides, our model captures more details than RDFNet on clothes stand. In row 3, cabinet in dark red is hard to recognize in RGB 
but with identifiable features in HHA. Improper fusion of RGB and HHA leads to erroneous semantics for this area (column $3$). 
While our model pays more attention to HHA in this area to achieve more precise results. 

\subsection{Comparing with State-of-the-arts}

\noindent \textbf{NYU Depth V2.}
Results are shown in Table \ref{sota-nyu}. Our model achieves leading performance. On the consideration of a fair comparison to ~\cite{pap,hu2019acnet,padnet} that utilize ResNet-50 as backbone, we also use same backbone and achieve $51.3\%$ \emph{mIoU}, which is still better than these methods. Specifically, ~\cite{park2017rdfnet,hu2019acnet} try to use channel-wise attention or vanilla convolution to extract complementary feature, which are more implicit than our model in selecting valid feature from complementary information.
Besides, we can see that utilizing depth data as extra supervision (such as ~\cite{pap,padnet}) could make network more robust than general RGB-D methods that take both RGB and depth as input sources~\cite{park2017rdfnet,cheng2017locality,3dgnn}. However, our results demonstrate that once the input RGB-D information could be effectively recalibrated and aggregated, higher performance could be obtained.

\noindent \textbf{CityScapes.}
We achieve $81.7$\% \emph{mIoU} on validation set and $82.8$\% \emph{mIoU} on test set, which are both leading performances.  Table \ref{city-test} shows results on test set. We observe that due to serious noise of depth measurements in this dataset, most of previous RGB-D based methods even worse than RGB-based methods. However, our method effectively distills depth feature and extracts valid information in it and boosts the performance. Note that ~\cite{choi2020cars} is a contemporary work and we outperform them by $0.7\%$. 
We exclude the results of GSCNN~\cite{gatedscnn} for fair comparison, since it uses a stronger backbone WideResNet instead of ResNet-101. However, we still outperform GSCNN by $0.9\%$ mIoU on the validation set and achieve the same performance as it on test set.

\section{Conclusion}

In this work, we propose a cross-modality guided encoder along with SA-Gate and BMP modules to address two key challenges in RGB-D semantic segmentation,~\textit{i.e.,} the effective unified representation for different modalities and the robustness to low-quality depth source. Meanwhile, our proposed encoder can act as a plug-and-play module, which can be easily injected to current state-of-the-art RGB semantic segmentation frameworks to boost their performances.\\

\noindent \textbf{Acknowledgments:} This work is supported by the National Key Research and Development Program of China (2017YFB1002601, 2016QY02D0304), National Natural Science Foundation of China (61375022, 61403005, 61632003), Beijing Advanced Innovation Center for Intelligent Robots and Systems (2018IRS11), and PEK-SenseTime Joint Laboratory of Machine Vision.

\clearpage
%
%
\bibliographystyle{splncs04}
\bibliography{egbib}
\clearpage

\section*{Appendix}
\setcounter{section}{0}
\section{Introduction}

This supplementary material presents: $(1)$ more implementation details based on the main paper; $(2)$ additional experimental analysis and qualitative results of our approach on NYU Depth V2, CityScapes \textit{val} set and SUN-RGBD dataset.

\section{Implementation Details}
We use PyTorch framework to implement our experiments. We set batch size to 16 for all experiments. We adopt mini-batch SGD with momentum to train our model. The momentum is fixed as $0.9$ and the weight decay is set to $0.0005$. We employ a poly learning rate policy where the initial learning rate is multiplied by $(1-\frac{iter}{max\_iter})^{0.9}$.  

For NYU Depth V2, we randomly crop the image to $480 \times 480$ and train $800$ epochs with base learning rate set to $0.02$. We employ cross-entropy loss on both the final output and the intermediate feature map output from ResNet-101 block4, where the weight over the final loss is $1$ and the auxiliary loss is $0.2$. 

For SUN-RGBD, we randomly crop the image to $480 \times 480$ and train $80$ epochs with base learning rate set to $0.02$. Cross-entropy loss is used for the final output.

For CityScapes, we randomly crop the image to $800 \times 800$ and train $240$ epochs with base learning rate set to $0.04$. We use OHEM loss for better learning. For data augmentation, we use random horizontal flipping and random scaling with scale $\{0.5, 0.75, 1, 1.25, 1.5, 1.75\}$. When comparing with the state-of-the-art methods, we adopt flipping and multi-scale inference strategies as a test-time augmentation to boost the performance.

\section{Experimental Results}

Besides the results analyzed in the main paper, we also conduct experiments on CityScapes \textit{val} set and SUN-RGBD dataset to further verify the effectiveness and generalization ablity of our approach. Meanwhile, we conduct more ablation studies on NYU Depth V2 to verify the robustness of the proposed method.
\begin{table}[t]
\begin{center}
\caption{CityScapes \textit{val} set results in terms of mIoU metric. We also list the results of RGB-based methods for reference}
\label{sota-city}
\resizebox{0.9\columnwidth}{!}{
\setlength{\tabcolsep}{2mm}{
\begin{tabular}{lllc}
\toprule
\textbf{Method}& \textbf{Depth Data} & \textbf{Backbone} & \textbf{mIoU(\%) } \\
\hline
GSCNN ~\cite{gatedscnn} & &WideResNet-101& 80.8 \\
CCNet~\cite{huang2018ccnet} & &ResNet-101& 81.3 \\
DANet~\cite{danet} &&ResNet-101& 81.5 \\
ACFNet~\cite{zhang2019acfnet}&&ResNet-101 & 81.5 \\
\hline
PADNet  ~\cite{padnet} &$\makecell[c]{\surd}$&ResNet-50& 76.1 \\
Shu Kong $et\ al.$~\cite{kong2018recurrent}&$\makecell[c]{\surd}$&ResNet-101 & 79.1 \\
\hline
RGB baseline && ResNet-101&80.5 \\
RGB-D baseline &$\makecell[c]{\surd}$&ResNet-101 & 80.5 \\
Ours  &$\makecell[c]{\surd}$&ResNet-50 & \underline{80.7} \\		
Ours  &$\makecell[c]{\surd}$& ResNet-101&\textbf{81.7} \\
\bottomrule
\end{tabular}
}}
\end{center}
\end{table}

\subsection{Results on CityScapes}
\noindent \textbf{Comparison with State-of-the-art Methods on CityScapes \textit{Val} Set.}  
Tabel \ref{sota-city} shows the results on CityScapes \textit{val} set comparing with state-of-the-art RGB-D based methods. We also list the results of RGB based methods for reference. We see that from row $6$ and $7$, due to the serious noisy depth measurements on this out-door dataset, a simple multi-modal fusion mechanism (RGB-D baseline) can not help explore the strength of depth data to boost the performance compared with its single modality version (RGB baseline). However,  with our proposed SA-Gate and BMP strategy, our final model could filter the noisy features and aggregate the cross-modality features more effectively. Thus, the proposed approach could still gain  $1.2\%$ mIoU increase to baselines. On the other hand, we see that from row $4$ to row $9$, our method is more robust than state-of-the-art RGBD-based methods that predict depth value either as a second stage multi-modal input~\cite{padnet} or as a gating module for RGB-feature~\cite{kong2018recurrent}. Our final model achieves 2.6\% mIoU improvement compared with~\cite{kong2018recurrent} under the setting of ResNet-101 backbone and 4.6\% mIoU improvement compared with~\cite{padnet} under the setting of ResNet-50 backbone. Comparing with raw depth source, although the predicted one avoids noisy information from the raw data, it will lead to the loss of depth information due to the over-smooth predicted depth values between objects.  The results demonstrate the effectiveness of our cross-modality feature propagation and the potential superiority of directly forwarding the `raw' depth source to the network than prediction ones as long as the noisy information can be effectively suppressed and multi-modality information can be fully explored.

\begin{table}[]
\begin{center}
\caption{Segmentation results on SUN-RGBD test set}
\label{sota-sunrgbd}
\resizebox{0.6\columnwidth}{!}{
\setlength{\tabcolsep}{2mm}{
\begin{tabular}{lcc}
\toprule
\textbf{Method} & \textbf{mIoU(\%) }  & \textbf{Pixel Acc.(\%)} \\
\hline
 Depth-aware CNN~\cite{depthaware} & 42.0 & - \\
 Kong $et\ al.$~\cite{kong2018recurrent} & 45.1 & 80.3 \\
 3DGNN~\cite{3dgnn} & 45.9 & - \\
 RDF-152~\cite{park2017rdfnet} & 47.7 & 81.5\\
 CFN~\cite{cfn} & 48.1 & - \\
 ACNet~\cite{hu2019acnet} & 48.1 & - \\
 PAP~\cite{pap} & \textbf{50.5} & \textbf{83.8}\\
 \hline
RGB baseline  & 46.0 & 81.0\\
RGBD baseline & 47.5 & 81.5\\
Ours & \underline{49.4} & \underline{82.5} \\
\bottomrule
\end{tabular}
}}
\end{center}
\end{table}

\subsection{Results on SUN-RGBD}

We further perform experiments on SUN-RGBD dataset to evaluate the effectiveness of our method. SUN-RGBD dataset contains images from several different datasets. It has $37$ categories of objects and consists of $10335$ RGB-D images. There are $5285$ images for training and $5050$ images for testing. We keep all hyper-parameters the same as NYU Depth V2 except the number of epochs. 

Quantitative results are shown in Table \ref{sota-sunrgbd}. We outperform most of the state-of-the-art methods, while slightly lower than  PAP~\cite{pap} on this dataset. When compare with our baselines, our final model could boost RGB baseline by $3.4\%$ mIoU and RGBD baseline by $1.9\%$ mIoU.

\section{Robustness to Noisy Signals Existing in the Input}
As claimed in the main paper, one of our goal is to devise the module that could effectively suppress noisy signals existing in the input depth measurement and highlight its positive features. To further verify this, we add different level Gaussian noise (mean=0, std ranges from 10 to 120) to the input HHA map and take our final model and RGB-D baseline (dual-branch DeepLab V3+) as a comparison. Note that the value of the input HHA map ranges from 0 to 255, just like the RGB image. Experimental results are listed in Table \ref{robustness-nyu}. We do not use multi-scale inference strategy here.

From the table, we observe several interesting phenomena as follows. (1) When adding small Gaussian noise with std=10, the input HHA map does not change considerably and the performance of the baseline and our final model only drop a little. However, our model has a smaller decrease than the baseline, which illustrates our model is more robust. (2) When we add large Gaussian noise with std=40, the performance of the baseline decreases more quickly than our final model (-44.6\textperthousand VS -22.7\textperthousand). When adding Gaussian noise with std=120, the performance of baseline drops -72.8\textperthousand, while our final model only drops -28.0\textperthousand. We attribute the robustness of our method to the filtering and recalibration operation in the SA-Gate, which may adaptively and effectively filter the noise of the input HHA map. Besides, since there is no clean ground-truth depth information to explicitly supervise the acts of SA-Gate, maybe exploring more explicit constraints on modules like SA-Gate will further enhance the robustness of the network to noisy scenes. We leave this to future work.

\begin{table}[t]
\begin{center}
\caption{Robustness test on NYU Depth V2 test set. Std=x means we add Gaussian noise with mean 0 and std x to the input HHA map. Results are shown in mIoU(\%)}
\label{robustness-nyu}
\resizebox{\columnwidth}{!}{
\setlength{\tabcolsep}{2mm}{
\begin{tabular}{lccccc}
\toprule
\textbf{Method}& \textbf{No Noise} & \textbf{Std=10} & \textbf{Std=40} & \textbf{Std=80} & \textbf{Std=120}\\
\hline
RGB-D Base & 48.91 & 48.79(-2.5\textperthousand) & 46.73(-44.6\textperthousand) & 46.18(-55.8\textperthousand) & 45.35(-72.8\textperthousand) \\
Ours & 51.50 & 51.41(-1.7\textperthousand) & 50.33(-22.7\textperthousand) & 50.23(-24.7\textperthousand) & 50.06(-28.0\textperthousand) \\
\bottomrule
\end{tabular}
}}
\end{center}
\end{table}

\section{Does Filtering and Recalibration Help? }

\begin{figure}[t]
\centering
\includegraphics[width=0.7\textwidth]{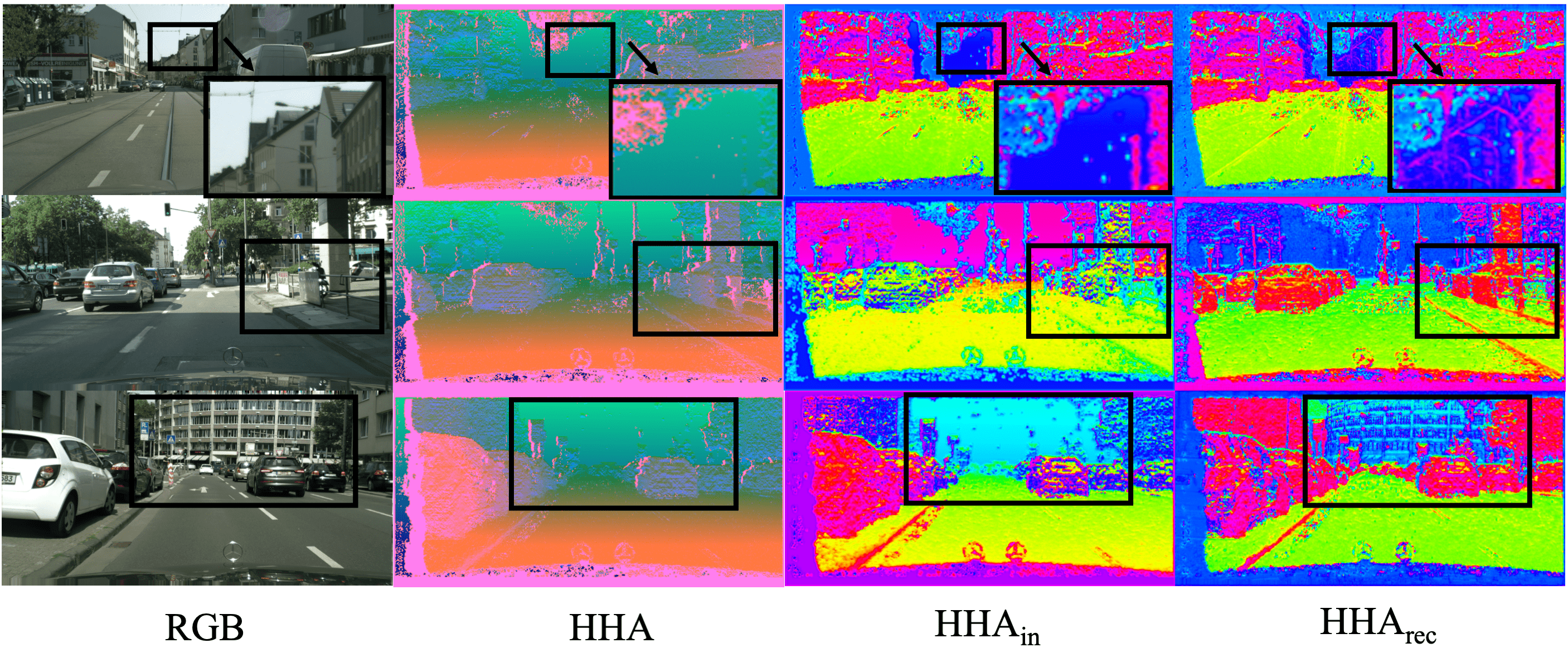} 
\caption{Visualization of depth feature before and after FSP on CityScapes val set}
\label{denoise-city-hha-supp}
\end{figure}

\begin{figure}[]
\centering
\includegraphics[width=0.7\textwidth]{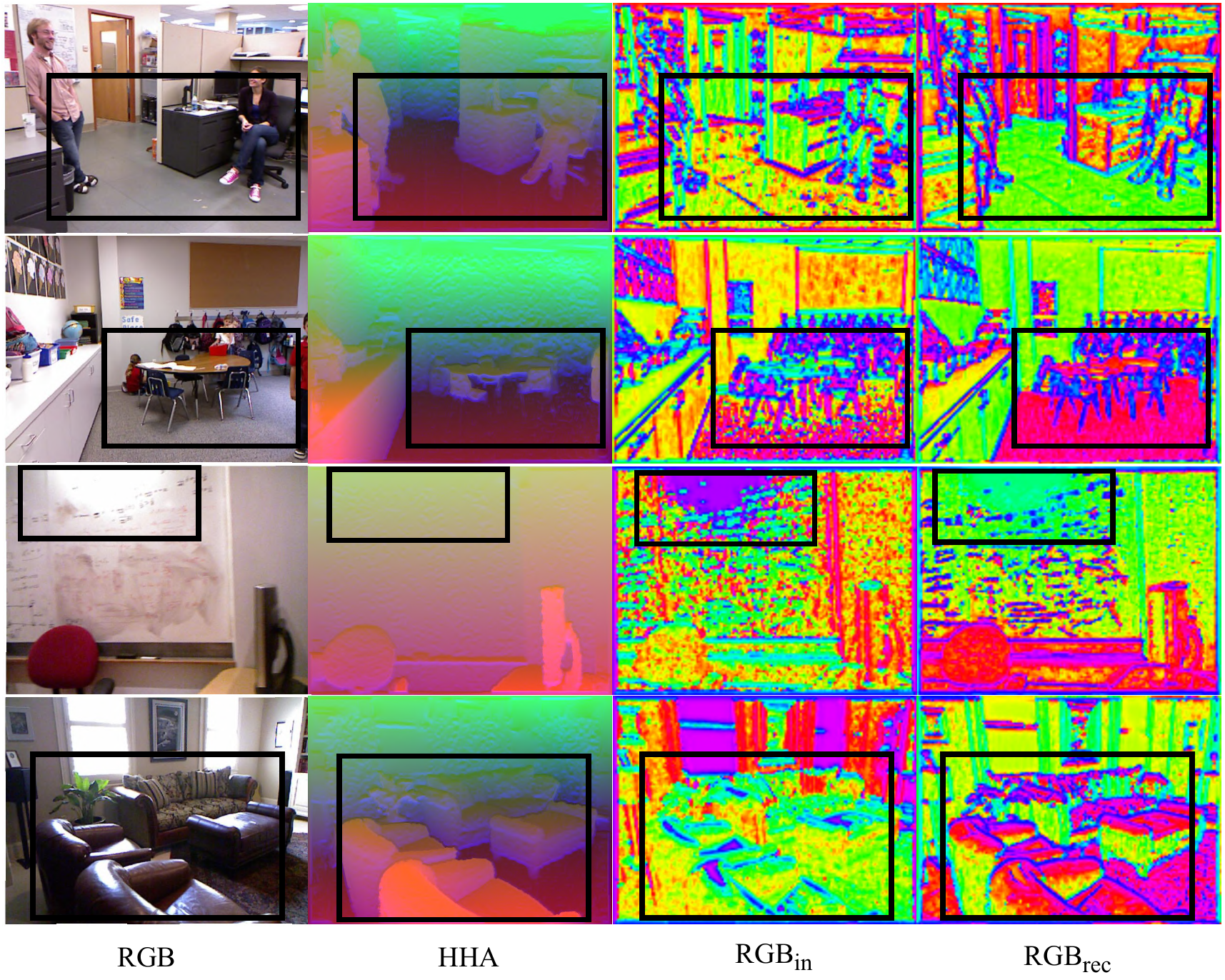} 
\caption{Visualization of RGB feature before and after FSP on NYU test set}
\label{denoise-nyu-rgb}
\end{figure}

 In Figure \ref{denoise-city-hha-supp}{\footnote{It has been shown in the main paper, but we list again here for more detailed explanations and discussions}}and Figure \ref{denoise-nyu-rgb}, we highlight several representative feature responses samples before and after \textit{Feature Separation Part} of proposed SA-Gate on both out-door and in-door datasets, to show how cross-modality feature filtering and recalibration can help refine primitive noisy single modality features in a more intuitive way.  We select the feature embedding computed by the first layer in our network and follow~\cite{zhuang2018relationnet} to compress the feature to three dimensions by the PCA and convert it to an RGB image for visualization. Note that the change of color does not directly relate to different feature responses. Instead, the color consistency inside objects indicates whether the module learns appropriate features.

We first visualize the response of HHA features on CityScapes $val$ set in Figure \ref{denoise-city-hha-supp}. 
In the first row, the streetlight is totally missing in the HHA image. After FSP, we can observe that the feature map shows a good response to the location of the streetlight. In the second row, the outline of the pole is more precise after FSP. In the third row, some objects which don't exist in the HHA images show up after FSP.

Different from out-door environments,  in-door scenes are more likely to decrease the validity of RGB modality due to the lighting and similar appearance of objects. Therefore,  we also visualize the response of RGB features on NYU Depth V2 test set in Figure \ref{denoise-nyu-rgb}.  In the first and the second rows, some unnecessary texture information is removed after FSP and we get a much smoother surface on the ground. In the third row, the effect of strong lighting is eliminated after FSP. In the fourth row, areas with inconspicuous contrast in the RGB image are enhanced after FSP.

In conclusion, the cross-modality feature filtering and recalibration in the proposed SA-Gate module could help make full use of the advantages of multi-modal data to supplement the missing signal and suppress unnecessary noisy feature responses.


\section{ Qualitative Results and Discussion}
 In this section, we present qualitative segmentation results of our method on in-door and out-door datasets, to show how cross-modality feature propagation could help semantic segmentation in various ways.

\begin{figure}[]
\centering
\includegraphics[width=0.9\textwidth]{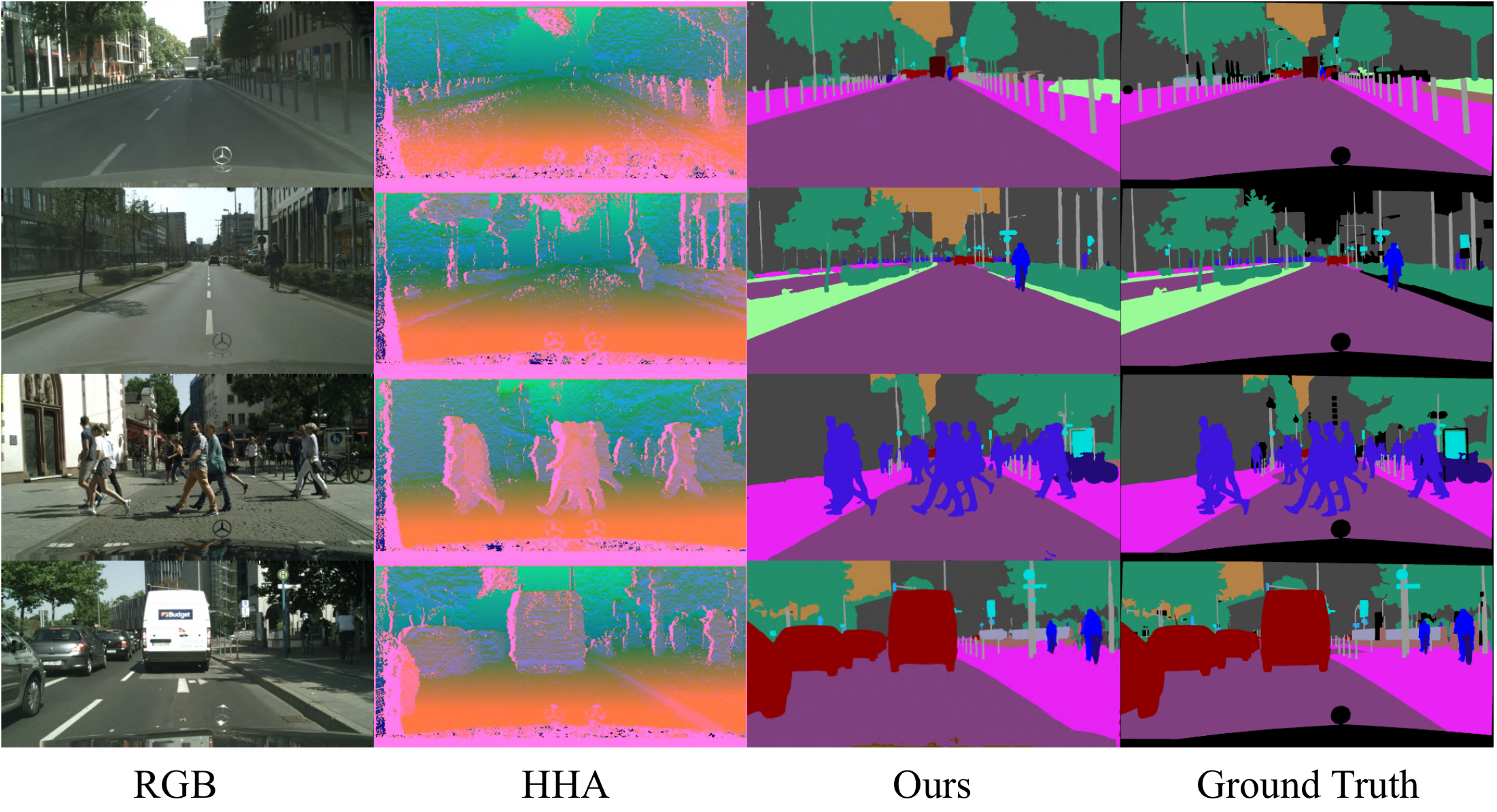}
\caption{Qualitative results on CityScapes \textit{val} set. Better viewed in color and zoom in}
\label{fig:results-city}
\end{figure}

Figure \ref{fig:results-city} shows some qualitative results on CityScapes. We observe that although the quality of depth source in CityScapes is very noisy and ambiguous, our model still achieve accurate segmentation results. Taking the first row as an example, the poles in the HHA image are indistinct, while our method successfully identifies the shape of the poles with the help of the RGB image. 

\begin{figure}[]
\centering
\includegraphics[width=0.9\textwidth]{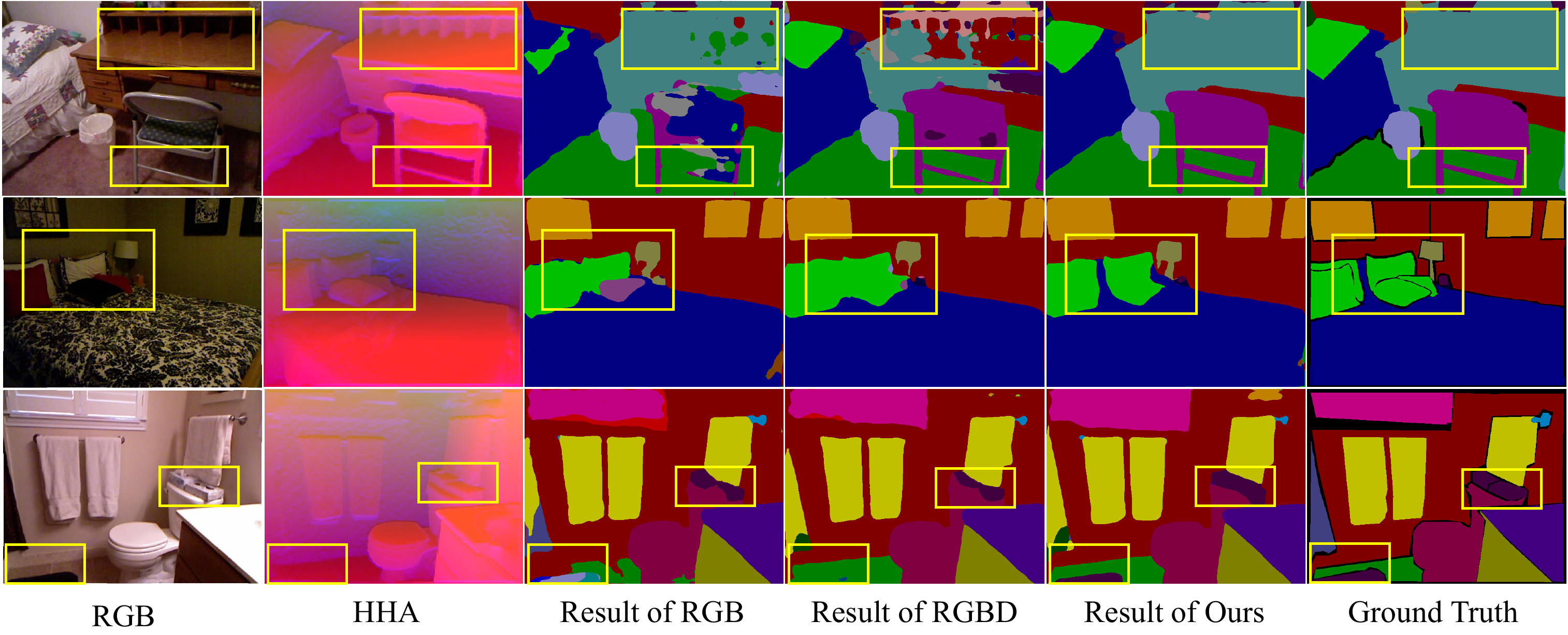}
\caption{\small{Results on NYU Depth V2 test set. From left to right: (1) RGB, (2) HHA, (3) result of RGB baseline, (4) result of RGB-D baseline, (5) result of ours, (6) groundtruth}}
\label{fig:results-nyu}
\end{figure}

 Figure \ref{fig:results-nyu} visualizes results on NYU Depth V2. In the first row, the desk is correctly identified in the RGB baseline, which is wrong in the RGB-D's. We can observe that part of the desk is misidentified as ~`wall', since it has the same orientation as the wall. The situation of the chair near the desk is just the opposite. Our method perfectly combines the characteristics of RGB and HHA to make both objects well recognized. In the second row, our method generates smoother object boundaries and gains better intra-class consistency. In the third row, the carpet in the lower-left corner is missing in the RGB-D baseline because it is attached to the ground, but it is well recognized with the proposed method.
 
\begin{figure}[t]
\centering
\includegraphics[width=0.88\textwidth]{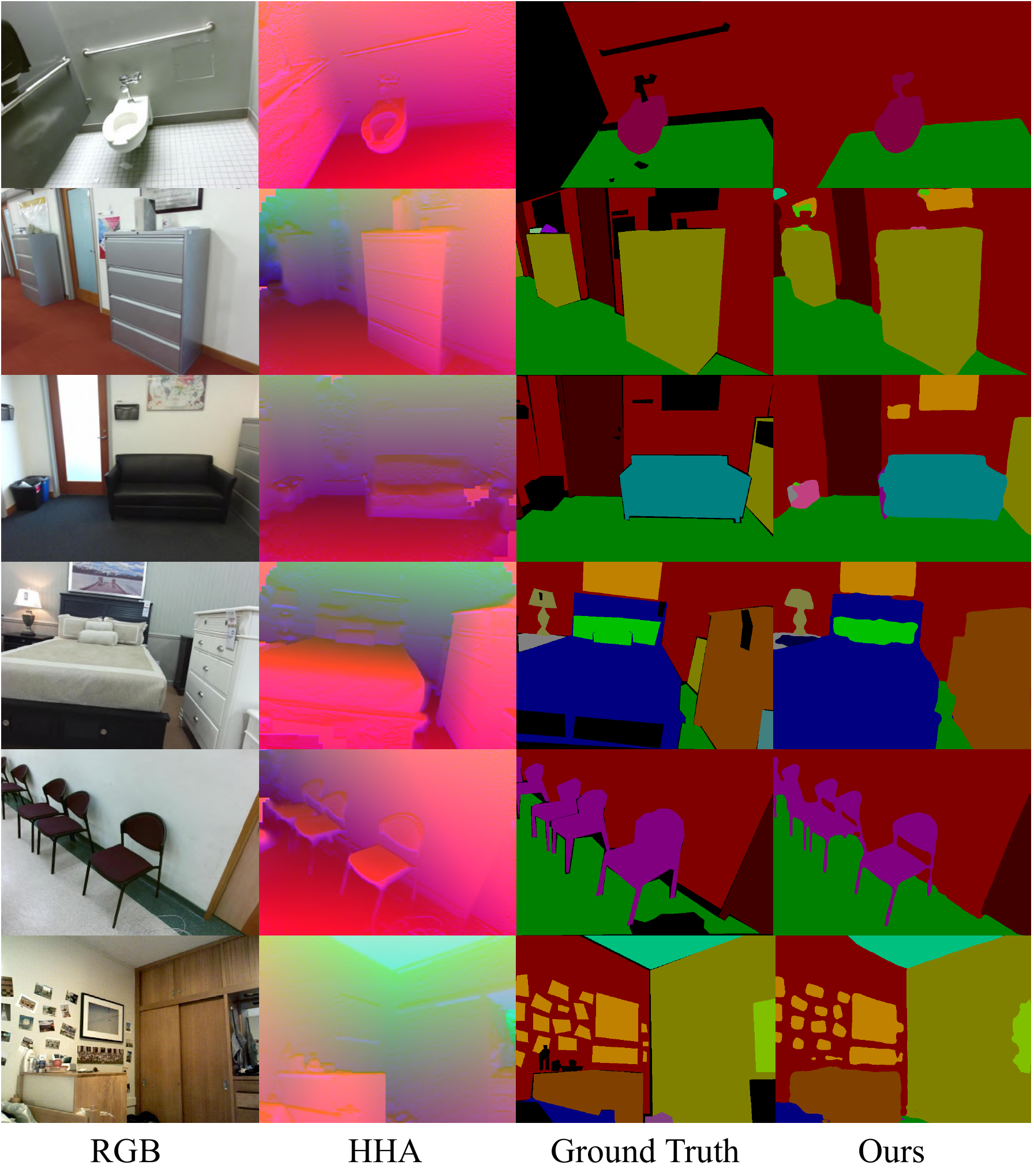} 
\caption{Qualitative segmentation examples on SUN-RGBD}
\label{sun-res}
\end{figure}

\begin{figure}[t]
\centering
\includegraphics[width=1\textwidth]{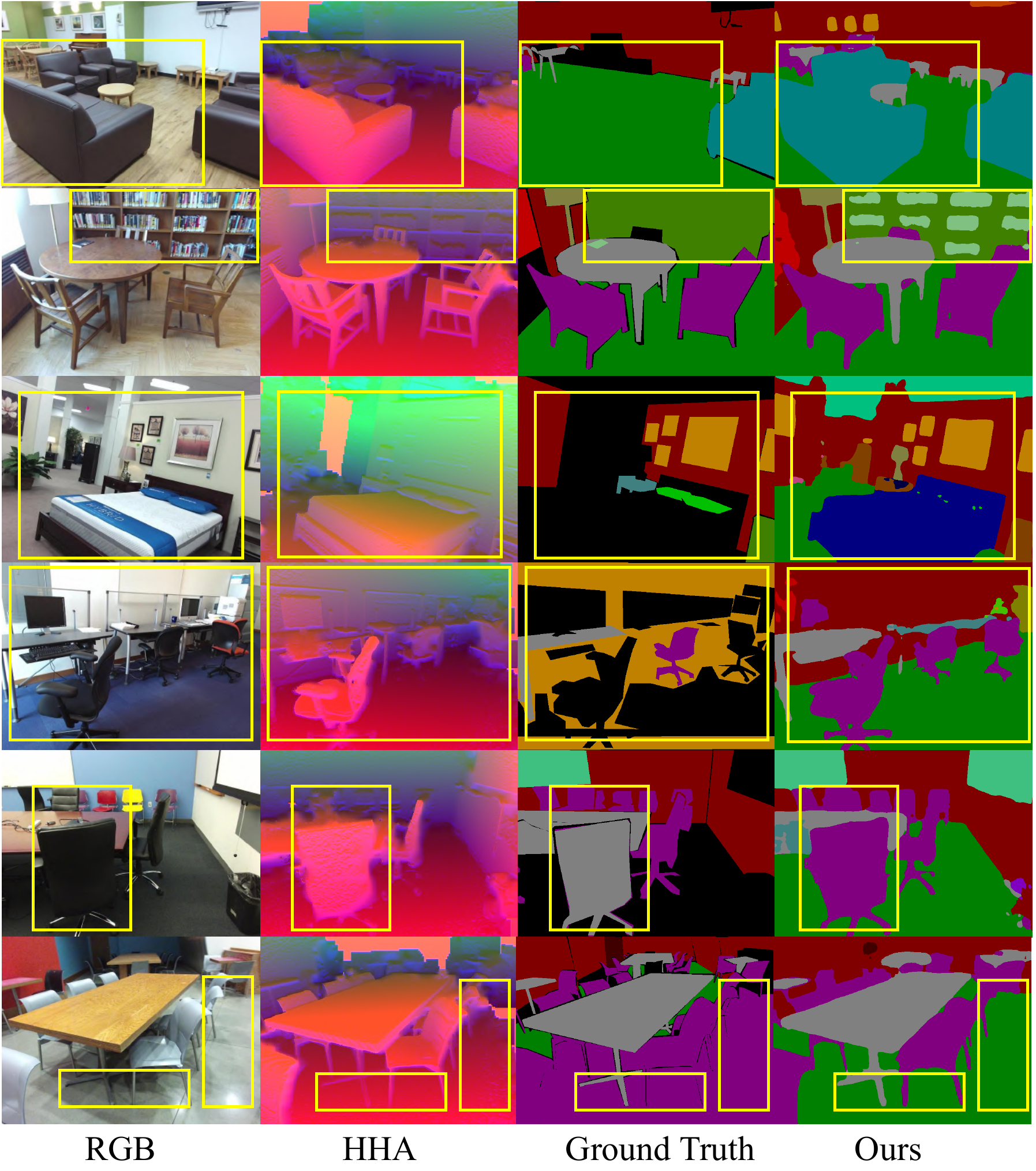} 
\caption{Some incorrect ground-truth labels on SUN-RGBD}
\label{sun-bad}
\end{figure}

Figure \ref{sun-res} shows our results on SUN-RGBD. We observe that our model handles the details very well and achieve satisfactory intra-class consistency and inter-class distinction. Figure~\ref{sun-bad} also reveals the strengths of our method. For example,  in the first row, the two sofas are missing in the ground truth. In the second row, books on the bookshelf are missing in the ground truth, which is due to coarse labeling. In the third and the fourth rows, many meaningful areas are set to invalid areas (marked as black). In the fifth row, the chair is mislabeled as a table. In the last row, a large region of the floor is mislabeled as chairs. However, our model recognizes these miss labeled objects correctly in the scene, since our method can make full use of complementary information in multi-modal data.

\clearpage
%
%
\bibliographystyle{splncs04}
\end{document}